\documentclass{article}

\usepackage{PRIMEarxiv}

\usepackage[utf8]{inputenc} % allow utf-8 input
\usepackage[T1]{fontenc}    % use 8-bit T1 fonts
\usepackage{hyperref}       % hyperlinks
\usepackage{url}            % simple URL typesetting
\usepackage{booktabs}       % professional-quality tables
\usepackage{amsmath,amsfonts}
\usepackage{nicefrac}       % compact symbols for 1/2, etc.
\usepackage{microtype}      % microtypography
\usepackage{lipsum}
\usepackage{fancyhdr}       % header
\usepackage{graphicx}       % graphics
\graphicspath{{media/}}     % organize your images and other figures under media/ folder

\usepackage[linesnumbered,lined,commentsnumbered, ruled]{algorithm2e} % For algorithms
\usepackage{multirow}
\usepackage{soul}
\usepackage{xcolor}
\usepackage{cite}
\usepackage{svg}

\title{Graph Convolutional Neural Networks with \\
Diverse Negative Samples via Decomposed Determinant Point Processes
}

\author{
  Wei Duan\\
  Australian Artificial Intelligence Institute\\
  University of Technology Sydney\\
  Australia\\
  \texttt{wei.duan@student.uts.edu.au} \\
  \And
  Junyu Xuan\\
  Australian Artificial Intelligence Institute\\
  University of Technology Sydney\\
  Australia\\
  \texttt{Junyu.Xuan@uts.edu.au} \\
   \And
  Maoying Qiao \\
  Australian Catholic University\\
  Australia\\
  \texttt{ maoying.qiao@acu.edu.au} \\
  \And
  Jie Lu\\
  Australian Artificial Intelligence Institute\\
  University of Technology Sydney\\
  Australia\\
  \texttt{Jie.Lu@uts.edu.au} \\
}

\begin{document}
\maketitle

\begin{abstract}
Graph convolutional networks (GCNs) have achieved great success in graph representation learning by extracting high-level features from nodes and their topology. Since GCNs generally follow a message-passing mechanism, each node aggregates information from its first-order neighbour to update its representation. As a result, the representations of nodes with edges between them should be positively correlated and thus can be considered positive samples. However, there are more non-neighbour nodes in the whole graph, which provide diverse and useful information for the representation update. Two non-adjacent nodes usually have different representations, which can be seen as negative samples. Besides the node representations, the structural information of the graph is also crucial for learning. In this paper, we used quality-diversity decomposition in determinant point processes (DPP) to obtain diverse negative samples. When defining a distribution on diverse subsets of all non-neighbouring nodes, we incorporate both graph structure information and node representations. Since the DPP sampling process requires matrix eigenvalue decomposition, we propose a new shortest-path-base method to improve computational efficiency. Finally, we incorporate the obtained negative samples into the graph convolution operation. The ideas are evaluated empirically in experiments on node classification tasks. These experiments show that the newly proposed methods not only improve the overall performance of standard representation learning but also significantly alleviate over-smoothing problems.
\end{abstract}

% keywords can be removed

\section{Introduction}
Convolutional neural networks (CNNs)\cite{lecun1995convolutional} learn potential features from large amounts of Euclidean data, such as text\cite{DBLP:journals/tkde/TrotzekKF20}, images\cite{duan2021refined}, music\cite{DBLP:conf/icassp/ChoiFSC17}, and video\cite{DBLP:journals/tnn/ChenWLXW18}, and achieve satisfactory performance on pattern recognition and data mining. In addition to Euclidean data, graphs, as non-Euclidean data, are powerful structures for modelling molecules, social networks, citation networks, traffic networks, etc.\cite{chakrabarti2006graph}. However, to lend learning power from the Euclidean space to graphs is not trivial. Learning with graphs requires effective representation of their data structure\cite{DBLP:conf/iclr/XuHLJ19}. Recently, graph neural networks (GNNs) have achieved great success in graph representation learning by extracting high-level features from nodes and their topology\cite{DBLP:journals/tnn/WuPCLZY21}. A wide variety of modifications of GNNs have been proposed for a specific task, for instance, recurrent graph neural networks \cite{DBLP:journals/corr/LiTBZ15,DBLP:conf/icml/DaiKDSS18}, graph convolutional neural networks \cite{DBLP:journals/corr/BrunaZSL13,DBLP:conf/iclr/KipfW17,hamilton2017inductive,DBLP:conf/iclr/XuHLJ19,9476188}, randomly wired graph neural networks\cite{9576073}, and spatio-temporal graph convolutional networks \cite{DBLP:conf/ijcai/YuYZ18,DBLP:conf/ijcai/WuPLJZ19}. 

\begin{figure}[!t]
\centering
\includegraphics[width=0.6\columnwidth]{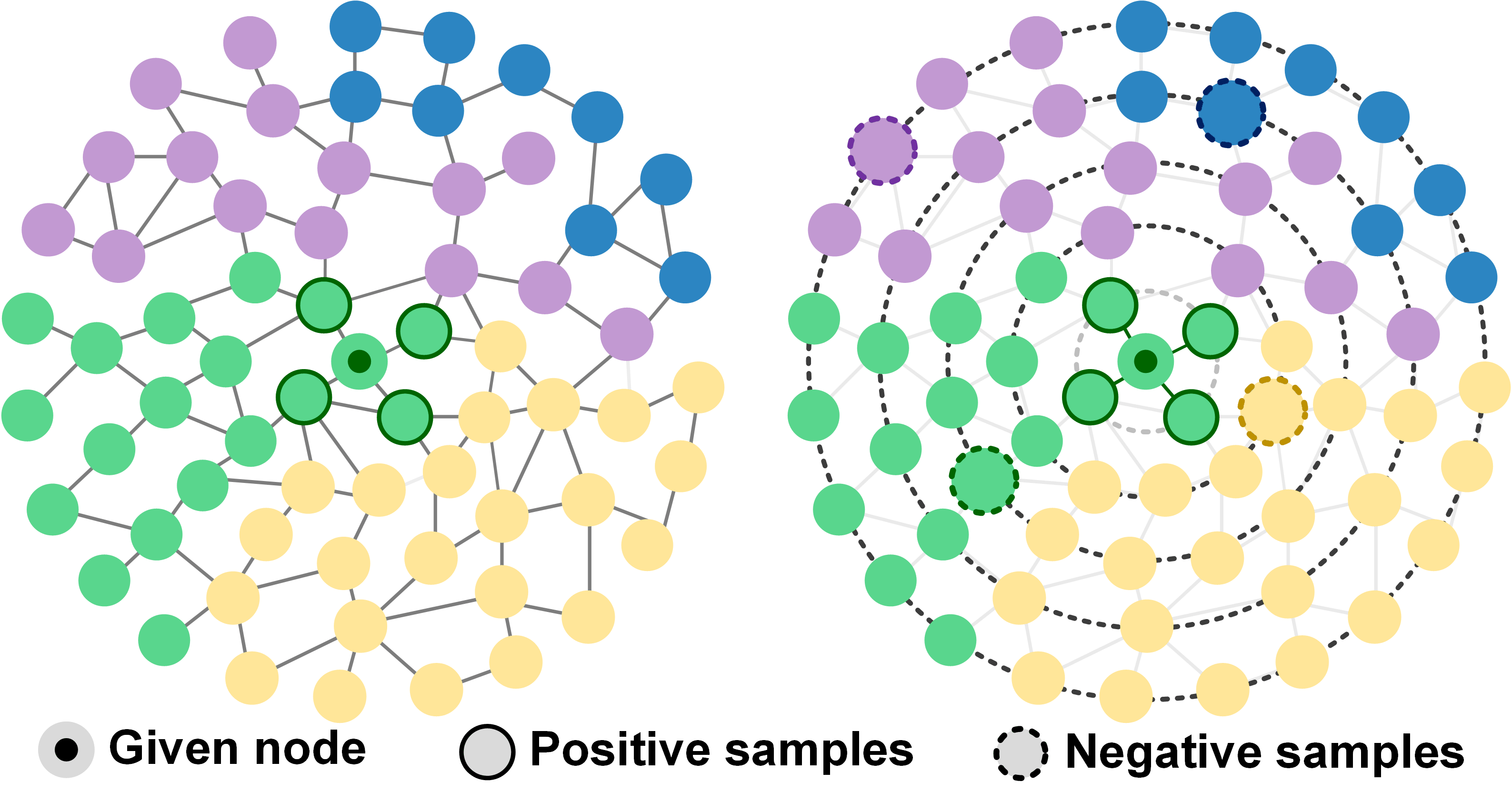} 
\caption{ Illustration of the motivation of this work. Left: For a given node, its first-order neighbours can be seen as positive samples.  Right: We compute the shortest path of the given node to other nodes. Different path lengths form concentric circles with different radii. Using DPP to select negative samples from different circles can get nodes for different clusters.}
\label{fig:Sapth illustration}
\end{figure}

One of the most representative GNNs is graph convolutional networks (GCNs), which introduce the
convolution operation to GNNs. Since GCNs generally follow a message passing mechanism\cite{geerts2021let}, each node aggregates information from its first-order neighbour to update its representation. As a result, the representations of nodes with edges between them should be positively correlated, and the first-order nearest neighbours of a node can be considered as positive samples for this node, as shown in the left part of Fig.\ref{fig:Sapth illustration}. As each node updates the representation only based on its neighbours, the representations of nodes become increasingly similar as the number of layers of the GCNs increases, leading to an annoying over-smoothing problem\cite{chen2020measuring}. 

In addition to neighbouring nodes, there are more non-neighbour nodes in the whole graph, which provide diverse and useful information for the representation update. They can further assist the model to better adapt to real-world data. Two non-adjacent nodes usually have different representations so that we can interpret them as negative samples. Although the importance of selecting appropriate negative samples in the graph is apparent, only a few studies have explored procedures to this end. The first method is to uniformly and randomly select some negative samples from non-neighbouring nodes\cite{kim2021find}, which will have a higher probability of picking the nodes inside the larger clusters but not the smaller ones. The second one is based on Monte Carlo chains \cite{yang2020understanding}. The third one is based on personalised PageRank\cite{ying2018graph}. These two methods only require the obtained negative samples to be uncorrelated with the positive samples while not obtaining a differentiated subset of negative samples. Thus, given the same number of negative samples, such negative samples could be redundant and cannot include as much information on the real world as possible. Hence, as a definition, the ideal negative samples should \textit{be diverse to include as much information on the whole graph as possible and without a large amount of redundant information.} Moreover, when selecting negative samples, besides the node representation, the structural information of the graph is also crucial for learning. Although some nodes have similar characteristics, they may have different positions in the graph and different topologies around the nodes, which leads to different effects on the network during the message passing. Therefore, when modelling the sample, using both node features and graph structure information will more fully exploit the structural characteristics of graph data.

In this paper, we first use the determinant point process (DPP) to obtain diverse negative samples. DPP is a powerful point process. The essential characteristic of DPP is that each item in the selected subset is negatively correlated, and a more diverse subset has a higher probability\cite{kulesza2012determinantal}. When defining a distribution on diverse subsets of all non-neighbouring nodes for a given node, we further utilize quality-diversity decomposition, which is a major fundamental technique of DPP, to incorporate graph structure information and node representations. Since the DPP sampling process requires matrix eigenvalue decomposition, if all nodes $N$ in the graph are used as a candidate set for the negative sample, the computational cost can be as high as $O(N^3)$. To improve the computational efficiency of this algorithm, we propose a new shortest-path-base method to reduce the size of the candidate set. For a given node, we compute its shortest path to other points and collect the endpoints of different path lengths with their first-order neighbours as the negative candidate set. The idea is that the shortest-path-based approximated method of a given node has the chance to select nodes for all the different clusters in the graph so the collection of all local diverse samples is able to better approximate all the information of the real world, as shown in the right part of Fig. \ref{fig:Sapth illustration}. Using this method, the computational cost is approximately  $O((\overline{\text{pa}} \cdot \overline{\text{deg}})^3)$ where $\overline{\text{pa}} \ll N$ is the path length (normally smaller than the diameter of the graph) and $\overline{\text{deg}} \ll N$ is the average degree of the graph. As an example, consider a Citeseer graph\cite{sen2008collective} with 3,327 nodes and $\overline{\text{deg}} = 2.74$. If using the setting of the experiment shown below that $\overline{\text{pa}} = 5$, we get $O(N^3) = 3.6e10 \gg 2571 = O((\overline{\text{pa}} \cdot \overline{\text{deg}})^3)$.
Finally, we incorporate the obtained negative samples into the graph convolution operation. The ideas are evaluated empirically in experiments on node classification tasks. These experiments show that the new proposed methods achieve superior results.

The contributions of our study are summarised as follows:
\begin{itemize}
\item A new negative sampling method is developed based on a quality-diversity decomposition in the determinant point process, which not only considers the feature representation of the node but also includes the graph structural information;
\item A shortest-path-base negative sampling method is developed to select nodes for different clusters, which can better approximate all the information of the whole graph. % real world.

\item Diverse negative samples are used to boost graph convolutional neural networks, which improve the performance of the node classification task and alleviate the over-smoothing problem.
\end{itemize}

\begin{table}[!t]
  \caption{Key notations}
     \begin{center}
      \begin{tabular}{c|p{0.75\columnwidth}}
        \hline
         Symbol & Meaning \\
         \hline 
         $G$ & a graph \\
         $G_n$ & the node set of graph $G$\\
         $\mathcal{Y}$ & a ground set \\
         $Y$ & a node subset\\
         $k$ & the number of negative samples of a given node in a graph\\
         $\mathcal{N}(i)$ &  neighbours (positive samples) of node $i$ \\
         $\overline{\mathcal{N}}(i)$ & negative samples of node $i$ \\
         $x_i^{(l)}$ & the representation of node $i$ at layer $l$ \\
         $\text{deg}(i)$ & the degree of node $i$ \\
         $\mathbb{L}$ & the $\mathbb{L}$-ensemble of DPP\\
         $\lambda$ & the eigenvalues of $\mathbb{L}$ \\
         $e_k^{|\mathcal{Y}|}$ & the $k^{th}$ elementary symmetric polynomial on eigenvalues $\lambda_1,\lambda_2,\dots, \lambda_{|\mathcal{Y}|}$ of $\mathbb{L}$ \\
         $\boldsymbol{v}$  & the eigenvectors of the $\mathbb{L}$-ensemble\\
         $ V$ & the set $\boldsymbol{v}$ \\
         \hline
      \end{tabular}
     \end{center}
  \label{tab:notation}
 \end{table}

\section{Preliminaries}
\label{sec:preliminary}

This section briefly introduces the basic graph convolutional neural networks using message passing and determinantal point processes. The key notations are given in Tab.\ref{tab:notation}.

\subsection{Graph convolutional neural networks (GCNs)}

GCNs\cite{DBLP:conf/iclr/KipfW17}  generalize the traditional concept of convolution from Euclidean data to graph data, which means sharing weights for nodes within a layer. The convolution process of GCNs can be regarded as a message-passing mechanism\cite{geerts2021let}, where each node receives information from its neighbours to generate its new representation. The vector form of a layer in GCNs is defined as

\begin{equation}
	{x_i}^{(l)} =\sum_{j \in \mathcal{N}_i \cup \{i\}}\frac{1}{\sqrt{\text{deg}(i)}\cdot\sqrt{\text{deg}(j)}}\big(\Theta^{(l)} \cdot x_j^{(l-1)}\big),
\label{eq:gcn}	
\end{equation} 
where $x_i^{(l)}$ is the representation of node $i$ at layer $l$, $\mathcal{N}(i)$ is the neighbours of node $i$ (i.e., positive samples), $\text{deg}(i)$ is the degree of node $i$, and $\Theta^{(l)}$ is a trainable weight. The goal of this equation is to aggregate the weighted representations of all neighbours. 

\subsection{Determinantal point processes (DPP)}

A point process $\mathcal{P}$ on a ground set $\mathcal{Y}$ is a probability measure over "point patterns", which are finite subsets of $\mathcal{Y}$, and $\mathcal{P}$ captures these correlations\cite{kulesza2012determinantal}. Given a ground set $\mathcal{Y}=\{1, 2, \dots, |\mathcal{Y}|\}$, a determinantal point process $\mathcal{P}$ is a probability measure on all possible subsets of $\mathcal{Y}$ with size $2^{|\mathcal{Y}|}$. For every $Y \subseteq \mathcal{Y}$, a DPP \cite{hough2009zeros} defined via an $\mathbb{L}$-ensemble is formulated as
\begin{align}
\mathcal{P}_{\mathbb{L}}(Y) = \frac{\det(\mathbb{L}_Y)}{\det(\mathbb{L} \pm I)},
\end{align}
where $\det(\cdot)$ denotes the determinant of a given matrix, $\mathbb{L}$ is a real and symmetric $|\mathcal{Y}| \times |\mathcal{Y}|$ matrix indexed by the elements of $\mathcal{Y}$, and $\det(\mathbb{L}+I)$ is a normalisation term that is constant once the ground dataset $\mathcal{Y}$ is fixed. Determinants have an intuitive geometric interpretation. Given a Gram decomposition $\mathbb{L}_Y = \bar{\mathbb{L}}^T \bar{\mathbb{L}}$, a determinantal operator can be interpreted geometrically as
\begin{align} 
\mathcal{P}_{\mathbb{L}}(Y) \propto \det(\mathbb{L}_Y) = \text{vol}^2 \left(\{\bar{\mathbb{L}}_i\}_{i\in Y} \right),
\end{align}
where the right-hand side is the squared volume of the parallelepiped spanned by the columns in $\bar{\mathbb{L}}$ corresponding to the elements in $Y$. Intuitively, to get a parallelepiped of greater volume, the columns should be as repulsive as possible to each other. Hence, DPP assigns a higher probability to a subset of $Y$ whose elements span a greater volume.

One important variant of DPP is $k$-DPP \cite{kulesza2012determinantal}. $k$-DPP measures only $k$-sized subsets of $\mathcal{Y}$ rather than all of them including an empty subset. It is formally defined as
\begin{align}
\mathcal{P}_{\mathbb{L}}(Y) = \frac{\det(\mathbb{L}_Y)}{e_k^{|\mathcal{Y}|}},
\label{eq:kdpp}
\end{align}
with the cardinality of subset $Y$ being a fixed size $k$, i.e., $|Y|=k$.
$e_k^{|\mathcal{Y}|}$ is the $k^{th}$ elementary symmetric polynomial on eigenvalues $\lambda_1,\lambda_2,\dots, \lambda_{|\mathcal{Y}|}$ of $\mathbb{L}$, i.e., $e_k(\lambda_1,\lambda_2,\dots, \lambda_{|\mathcal{Y}|})$.

Both DPP and $k$-DPP can be used to sample a diverse subset, and both have been well studied in the machine learning area \cite{kulesza2012determinantal}. The popularity of DPP (and also of $k$-DPP) for modelling diversity is because of its significant modelling power.

\section{Related Work}
\label{sec:related-work}
\subsection{GCN and its variants}

Motivated by the achievements of convolutional neural networks (CNNs), many GCN approaches have been actively studied to model graph data. These are generally categorized into two branches: spectral-based and spatial-based. Bruna et al. \cite{DBLP:journals/corr/BrunaZSL13} first created a graph convolution based on spectral graph theory. Although their paper was conceptually important, the graph Fourier transform is computationally
expensive, which prevented it from being a genuinely useful tool. Since then, a growing number of enhancements, extensions, and approximations of spectral-based GCNs have been made to overcome these flaws. Based on these, Kipf and Welling \cite{DBLP:conf/iclr/KipfW17} proposed GCNs, which is a localized first-order approximation of spectral graph convolutions as a generalised method for semi-supervised learning on graph-structured data. The matrix form of a layer in GCNs is defined as
 \begin{equation}\label{GCN_MATIX}
    X^{(k)} = \text{ReLU} \Big( \hat{A} X^{(k-1)} W^{(k-1)} \Big),
\end{equation}
\noindent
where $\hat{A} = \tilde{D}^{-\frac{1}{2}}\tilde{A}\tilde{D}^{-\frac{1}{2}}$,  $\tilde{A} = A + I_N$ and $\tilde{D}_{ii} = \sum_{j} \tilde{A}_{ij} $.
Their model acquires implicit representations, encodes the region graph structure and node attributes, and expands linearly regarding the number of graph edges. In addition to using GCNs for graphs in which nodes are only one type, many spectral-based works also focus on the heterogeneous graph, which contains multiple types of nodes and edges\cite{zhu2020hgcn,yang2021interpretable,li2021higher}.

A representative spatial-based work is GraphSAGE \cite{hamilton2017inductive}, which is a general framework for generating node embedding by sampling and aggregating features from the neighbourhood of a node. To theoretically analyze the representational power of GNNs, Xu et al. \cite{DBLP:conf/iclr/XuHLJ19} formally characterised how expressive different GNNs variants are at learning to represent different graph structures based on the graph isomorphism test \cite{weisfeiler1968reduction}. They proposed that, since modern GNNs follow a neighbourhood aggregation strategy, the network at the $k$-th layer can be summarised formally in two steps AGGREGATE and COMBINE
\begin{equation}
\begin{aligned}
    {a_{v}^{(k)}} &= \text{AGGREGATE}^{(k)}(\{{x_{u}^{(k-1)}}, \forall u \in \mathcal{N}(v) \}),\\
	{x_{v}^k} &=  \text{COMBINE}^{(k)}\big({x_{v}^{(k-1)}}, {a_{v}^{(k)}}) \big),
\end{aligned}
\end{equation}
\noindent
where $x_v^{(k)}$ is the feature vector of node $v$ in the $k^{th}$ layer.  In this way, the AGGREGATE and COMBINE in Eq.\ref{GCN_MATIX} will be defined as
\begin{equation}\label{eq:GCN}
    {x_{v}^{(k)}} \! =\! \text{ReLU} \! \Big(\! W\! \cdot \text{MEAN} \Big\{ {x_{u}^{(k-1)}},\forall u \in \mathcal{N}(v) \cup \{v\} \Big\}\Big),
\end{equation}
\noindent
where $W$ is a learnable weight matrix. AGGREGATE in GraphSAGE \cite{hamilton2017inductive} is defined as
\begin{equation}
    {a_{v}^{(k)}} = \max \left( \left\{ \text{ReLU}\left(W \cdot {x_{u}^{(k-1)}} \right), \forall u \in \mathcal{N}(v) \right\}\right),
\end{equation}
\noindent
where $MAX$ denotes an element-wise max-pooling.
In addition, they further proposed GINs\cite{DBLP:conf/iclr/XuHLJ19},
\begin{equation}
    {x_{v}^{(k)}} = \text{MLP}^{(k)} \Big( \big(1+\epsilon^{k} \big)\cdot {x_{v}^{(k-1)} +\sum_{u \in \mathcal{N}(v)} x_{u}^{(k-1)}} \Big),
\end{equation}
\noindent
which can distinguish different graph structures and capture dependencies between graph structures to achieve better classification results.

All the aforementioned GNNs use positive sampling when generating new node feature vectors because the neighbour aggregation on $j \in \mathcal{N}(i)$  leads to a high correlation between the central node and its neighbours. Studies using negative sampling in GNNs are relatively rare. Negative sampling was first used in natural language processing to simplify noise contrastive estimation\cite{mnih2013learning}, which can facilitate the training of word2vec\cite{mikolov2013distributed}. The only three works \mbox{\cite{ying2018graph,kim2021find,yang2020understanding}} that use negative sampling in GNNs cannot select good negative samples that include as much information on the real world as possible without much overlapping or redundant information. Moreover, the above approaches only use the results of negative sampling in the loss function, while the direct application of convolution operations remains unexplored.

\subsection{DPP and its applications}

Determinantal point processes (DPPs) \cite{hough2009zeros} are statistical models and provide probability measures over every configuration of subsets on data points. DPPs were first introduced to the machine learning area by Kulesza and Taskar \cite{kulesza2012determinantal}, and they have since been extended to include closed-form normalisation, marginalisation \cite{kulesza2012determinantal}, sampling \cite{kang2013fast}, dual
representation, maximising a posterior (MAP) \cite{gillenwater2012near}) and parameter learning \cite{affandi2014learning,gillenwater2014expectation}, and its structural \cite{gillenwater2012discovering} and Markov \cite{affandi2012markov} variants. This repulsive characteristic has been successfully applied to prior modelling in a variety of scenarios, such as clustering \cite{kang2013fast}, 
inhibition in neural spiking data \cite{snoek2013determinantal}, sequential labelling \cite{qiao2015diversified}, tweet timeline generation\cite{DBLP:conf/aaai/YaoFZWCX16}, document summarisation\cite{DBLP:conf/acl/ChoLFL19}, video summarization\cite{DBLP:conf/ijcai/ZhengL20} and so on.

\section{The Proposed Model}
\label{sec:our-proposed-model}

This section first introduces a method for obtaining good negative samples for a given node, including two negative sampling algorithms. We then integrate the algorithms with a GCN to obtain a new graph convolutional neural network.% boosted by diverse negative samples (D2GCN).

\subsection{DPP-based negative sampling}

Given a node, we believe its good negative samples should include as much information of the entire graph as possible and at the same time without much overlapping and redundant information. Hence, we propose to use DPP to generate negative samples, but applying DPP to different scenarios is not trivial and needs additional effort. 

The key to utilizing DPP is defining an $\mathbb{L}$-ensemble
which is  a real, symmetric matrix indexed by the elements of $Y$
\begin{align} 
\mathcal{P}_{\mathbb{L}}(Y) \propto \det(\mathbb{L}_Y).
\end{align}

In our previous study\cite{D2DCN}, only the node representations is calculated when we define an $\mathbb{L}$-ensemble
\begin{equation}
	\mathbb{L}_{G_{n\setminus i}}(j, j') = \exp \left( \cos \left(x_j, x_{j'} \right)-1 \right),
\label{eq:lensemble}	
\end{equation} 
where $G_{n\setminus i}$ denotes the node set of $G_n$ excluding node $i$, $j$ and $j'$ are two nodes within $G_{n\setminus i}$, and $\cos  (\cdot, \cdot)$ is the cosine similarity between two node representations $x$. 

\subsection{Decomposed DPP-based negative sampling}
Besides the node representations, the structural information of the graph is also crucial for learning. Although some nodes have similar features, their positions in the graph may be different, and the topology around the nodes is different, which leads to varying effects of these nodes on the network during the message-passing process. Therefore, when calculating the $\mathbb{L}$-ensemble, incorporating the structural information of the graph helps make fuller use of the characteristics of the structured data. We next describe how to incorporate the structural information of the graph and the features of the nodes into $\mathbb{L}$ using Decomposed DPP.

\begin{figure}[!t]
\centering
\includegraphics[width=0.6\columnwidth]{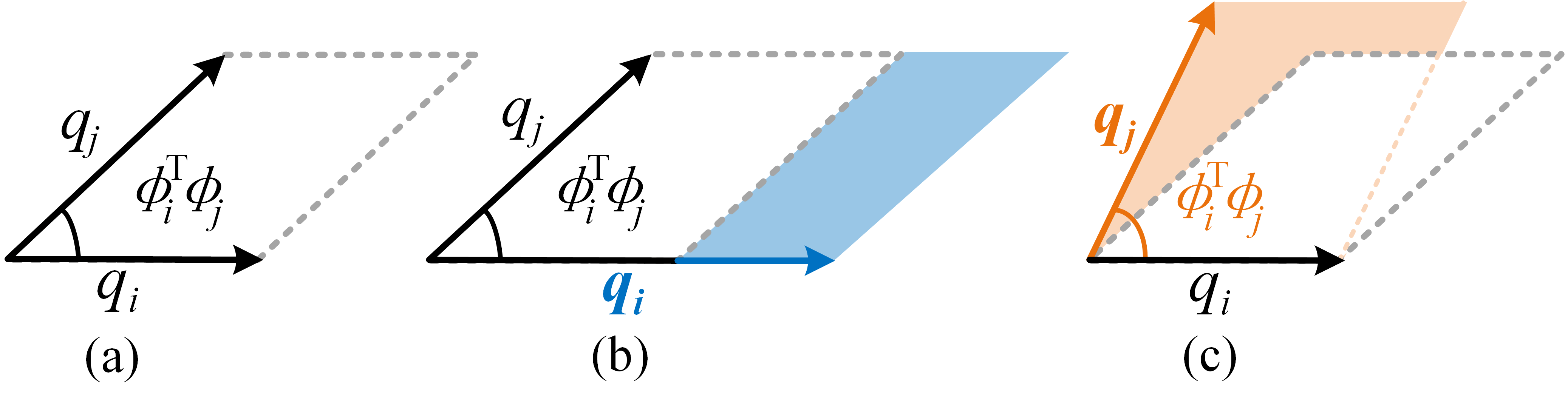} 
\caption{ A geometric view of Decomposed DPPs. (a) The probability of a subset $Y$ is the square of the volume spanned by $q_{i} \phi_{i}$ for $i\in Y$. (b) As item $i$'s quality $q_i$ increases, so do the probabilities
of sets containing item $i$. (c) As two items $i$ and $j$ become more dissimilar, the angle between two vectors and the probabilities of sets containing both $i$ and $j$ increase.}
\label{fig:decomposed illustration}
\end{figure}

One major fundamental technique of DPP is Quality-Diversity decomposition, which helps us balance the diversity against some underlying preferences for different items in $\mathcal{Y}$ \cite{kulesza2012determinantal}. Since $\mathbb{L}_Y$ can be written as a Gram matrix $\mathbb{L}_Y = \bar{\mathbb{L}}^T \bar{\mathbb{L}}$, each column $\bar{\mathbb{L}}$ is further written as the product of a \textbf{quality} term $q_i \in \mathbb{R}^+$ and a vector of normalized \textbf{diversity} features $\phi_i \in \mathbb{R}^D$,$	\left \| \phi_i \right\|=1 $.
As shown in Fig.\ref{fig:decomposed illustration}(a), the probability of a subset $Y$ is the square of the volume spanned by $q_{i} \phi_{i}$ for $i\in Y$. The $\mathbb{L}_Y$  now becomes as
\begin{align}
\mathbb{L}_{ij} = q_i\phi_{i}^{T}\phi_{j} q_j
\label{eq:decomDpp},
\end{align}
where $q_i \in \mathbb{R}^+$ measures the "goodness" of item i, and $\phi_{i}^{T}\phi_{j} \in [-1,1]$ measures the similarity between items i and j. We next describe how to incorporate the structural information of the graph and the features of the nodes into \textbf{quality term} and \textbf{diversity term}, respectively.

For all nodes $V$ in graph $G$, we first use a semi-synchronous label propagation (LPA) method\cite{DBLP:journals/corr/abs-1103-4550} to divide the nodes into $K$ communities as
\begin{equation}
    V = \{C_{k}\}_{k=1}^{K},
\label{eq:getcommunity}	
\end{equation} 
where $C_{k}$ denotes the set of nodes for each community.
Using the node representations $x_i$, the features of each community $a_k$ can be calculated as
\begin{equation}
    a_{k} =\frac{\sum_{i \in C_k} x_i}{\left|C_{k}\right|}, 
\label{eq:communityfea}	
\end{equation}
where $\left|C_{k}\right|$ is the number of nodes in $C_{k}$. 

For a given node $i$,  its negative samples will be selected from a candidate set $S_i$. The features of each candidate set $S_i$ are calculated as
\begin{equation}
    b_{i} =\frac{\sum_{j \in S_i} x_j}{\left|S_{i}\right|}, 
\label{eq:condidatefea}	
\end{equation}
where $\left|S_{i}\right|$ is the number of nodes in $S_{i}$. 
We first define the \textbf{quality term} as
\begin{equation}
\begin{aligned}
    &q_{i,j} = \cos(a_{i},b_{i})\otimes  \cos(a_{i},a_{j}),
    \\
    &{q_{i,j^{\prime}}} = \cos(a_{i},b_{i})\otimes  \cos(a_{i},{a_{j^{\prime}}}),
\end{aligned}
\label{eq:quality}	
\end{equation}
where $j,j^{'} \in S_i$, $a_{i}, a_{j},{a_{j^{\prime}}}$ represents the feature expression of the node belonging to its community, $b_i$ denotes the features of the candidate set $S_i$, and the  $\otimes$ means point-wise product. Then the \textbf{diversity term} is defined as
\begin{equation}
\begin{aligned}
   {\phi_{j}^{T}\phi_{j^{\prime}}}= & \cos({x_{j}},{a_{j^{\prime}}}) \cos({a_{j}},{x_{j^{\prime}}}) \\
    & \otimes \exp{( \cos(({x_j},{x_{j^{\prime}}})-1))},
\end{aligned}
\label{eq:diversity}	
\end{equation}
where $a_{j}, a_{{j^{\prime}}}$ represents the feature expression of the node belonging to its community. Put Eq.\ref{eq:quality}
and Eq.\ref{eq:diversity} into Eq.\ref{eq:decomDpp},we have

\begin{equation}
\mathbb{L}_{i,j,j^{\prime}} = q_{i,j} \phi_j^T \phi_{j^{\prime}} q_{i,j^{\prime}},
\label{eq:finalDPP}
\end{equation}
where $i$ is the given node, $j,{j^{\prime}} \in S_i$.

% Combining the node representations and community features,
For simplicity, we take the given node $i$ and candidate node $j$ to further illustrate our idea. The \textbf{quality term} $q_{ij}$ measures the "goodness" of items $j$. Specifically, in Eq.\ref{eq:quality}, $\cos(a_{i},b_{i})$ measures the similarity between the community feature of node $i$ and its candidate set, which means we want the candidate nodes in $S_i$ to be sufficiently different from the given node $i$ in terms of community feature. Moreover, $\cos(a_{i}, a_{j})$ measures the similarity between the communities of given node $i$ and candidate node $j$, which means the given node should not be in the community where the candidate nodes are located. Generally, Eq.\ref{eq:quality} helps us to obtain a higher quality candidate set $S_i$. Fig.\ref{fig:decomposed illustration}(b) makes clear that as item $i$'s quality $q_i$ increases, so do the probabilities of sets containing item $i$. As for the \textbf{diversity term}, the upper part of Eq.\ref{eq:diversity} measures the similarity between the node representation of $j$ and the community feature of $j^{\prime}$, and vice versa. In this way, two nodes that belong to different communities are selected to reduce redundant information. The lower part of Eq.\ref{eq:diversity}  measures the similarity between two node features $x_j$ and $x_{j^{\prime}}$ because when defining the distance between nodes, nodes with similar information are expected to be further apart. Generally, Eq.\ref{eq:diversity} helps us to obtain diverse nodes from the candidate set $S_i$. Fig.\ref{fig:decomposed illustration}(c) shows that as two items $i$ and $j$ become more dissimilar, the angle between two vectors and the probabilities of sets containing both $i$ and $j$ increase. The reason we chose cosine similarity is because the following prediction (label and link) is usually based on cosine similarity. It is worth noting that matrix $\mathbb{L}$  calculated using Eq.\ref{eq:finalDPP} is strictly symmetric. To ensure the positive definiteness of the $\mathbb{L}$, we add a small diagonal matrix $\epsilon I$ to $\mathbb{L}$ ($\epsilon = 0.01$). 

With this $\mathbb{L}$-ensemble, we obtain a distribution on all possible subsets of all nodes in the graph except for $i$,
\begin{equation}
\mathcal{P}_{\mathbb{L}}(Y) = \frac{\det(\mathbb{L}_{Y})}{\det(\mathbb{L}+I)},
\label{eq:py}
\end{equation} 
where $Y$ is a set of all subsets of $G_n$ and $\det (\cdot)$ is the matrix determinant operator. Compared with other ordinary probability distributions with the same support, this distribution has a unique property, namely the more diverse the subsets, the higher their probability values, and the easier it is to obtain a diverse set from this distribution by sampling. To ensure the scale of negative samples is similar to the positive samples, we use $k$-DPP to fix the number of negative samples as $k=|\mathcal{N}_i|+1$. Here, we use a sampling method based on eigendecomposition \cite{hough2006determinantal,kulesza2012determinantal}. Eq.\ref{eq:py} can be rewritten as the $k$-DPP distribution in terms of the corresponding DPP
\begin{equation}
\mathcal{P}_{\mathbb{L}}^{k}(Y) = \frac{1}{e_{k}^{|G_n|}}\det(\mathbb{L}+I)\mathcal{P}_{\mathbb{L}}(Y),
\label{eq:PLk}
\end{equation}
whenever $|Y| = k $ and $e_{k}^{|G|}$ denotes the $k$-th elementary symmetric polynomial. Following \cite{kulesza2012determinantal}, Eq.\ref{eq:PLk} is decomposed into elementary parts
\begin{equation}
\mathcal{P}_{\mathbb{L}}^{k}(Y) = \frac{1}{e_{k}^{|G_n|}}\sum_{|J|=k}\!\mathcal{P}^{V_{J}}(Y) \!\prod_{m\in J}\! {\lambda}_{m},
\label{eq:PLkelement}
\end{equation}
where $ V_{Y} $ denotes the set $\{\boldsymbol{v}_m\}_{m\in Y}$ and $\boldsymbol{v}_m$ and ${\lambda}_m $ are the eigenvectors and eigenvalues of the $\mathbb{L}$-ensemble, respectively. Based on Eq.\ref{eq:PLkelement}, for the self-contained reason, the complete process of sampling from $k$-DPP is given in Algorithm \ref{alg:sampling_DPP}. 

Note that compared with a sample, the mode of this distribution is a more rigorous output \cite{gillenwater2012near}, but it usually involves unacceptable complexity, so we use a sample rather than a mode here. The experiment evaluation shows that the sample achieves satisfactory results. Algorithm \ref{alg:dpp} shows a diverse negative sampling method based on DPP, but its computation cost is above the standard for a DPP on $G_n$ is $O(|G_n|^3)$ for each node. This totals an excessive $O(|G_n|^4)$ for all nodes! Although Algorithm \ref{alg:dpp} is able to explore the whole graph and find the best diverse negative samples, the large number of candidates makes it an impractical solution for even a moderately sized graph. Hence, we propose following the approximate heuristic method.

\begin{algorithm}[!t]
\caption{Obtain a sample from $k$-DPP \cite{kulesza2012determinantal}}
\label{alg:sampling_DPP}
\SetAlgoLined
\KwIn{size $k$, a DPP distribution $\mathcal{P}_{\mathbb{L}}(Y)$ on a set with size $V$}
\KwOut{$Y$ with size $|Y|=k$}
Eigen-decompose $\mathbb{L}$ to obtain eigenvalues $\lambda_1, \dots, \lambda_V$\; 
Iteratively calculate all elementary symmetric polynomials $e_l^v$ for $l=0, \dots, k$ and $v=0, \dots, V$ following \cite{kulesza2012determinantal}\;
$J \leftarrow \emptyset$\;
$l \leftarrow k$\;
\For{$v=V,\dots,1$}{
    \If{$l=0$}{break\
    }
    \If{$\mu \sim U[0,1] < \lambda_v \frac{e_{l-1}^{v-1}}{e_l^v}$}{
    $J \leftarrow J \cup \{v\}$\;
    $l \leftarrow l - 1$\;
    }
}
$V \leftarrow \left\{\boldsymbol{v}_n\right\}_{n \in J}$\;
$Y \leftarrow \emptyset$\;
\While{$|V|> 0 $}{
    Select $i$ from $\mathcal{Y}$ with $\operatorname{Pr}(i)=\frac{1}{|V|} \sum_{\boldsymbol{v} \in V}\left(\boldsymbol{v}^{\top} \boldsymbol{e}_i\right)^2$\;
    $Y \leftarrow Y \cup i$\;
    $V \leftarrow V_{\perp}$, an orthonormal basis for the subspace of $V$ orthogonal to $\boldsymbol{e}_i$\;
}
\end{algorithm}

\begin{algorithm}[!t]
\caption{Diverse negative sampling}
\label{alg:dpp}
\SetAlgoLined
\KwIn{A graph $G$}
\KwOut{$\overline{\mathcal{N}}(i)$ for all $i \in G$}
Let $\overline{\mathcal{N}} = \bf{0}$\;
Use label propagation method\cite{DBLP:journals/corr/abs-1103-4550} to divide the nodes into $K$ communities\;
\For{each node $i$}{
    Compute community feature $cf_i$ using Eq.\ref{eq:communityfea}\;
    Compute candidate set feature $sf_i$ using Eq.\ref{eq:condidatefea}\;
    Compute $\mathbb{L}_{G}$ using Eq.\ref{eq:finalDPP}\;
    Define a distribution on all possible subsets $\mathcal{P}_{\mathbb{L}}(Y_{G})$ using Eq.\ref{eq:py}\;
    Obtain a sample of $\mathcal{P}_{\mathbb{L}}(Y_{G})$ using Algorithm \ref{alg:sampling_DPP}\;
    Save nodes in the sample as $\overline{\mathcal{N}}(i)$\;
}
\end{algorithm}

\subsection{Shortest path based approximated method}

\begin{figure}[!t]
\centering
\includegraphics[width=0.4\columnwidth]{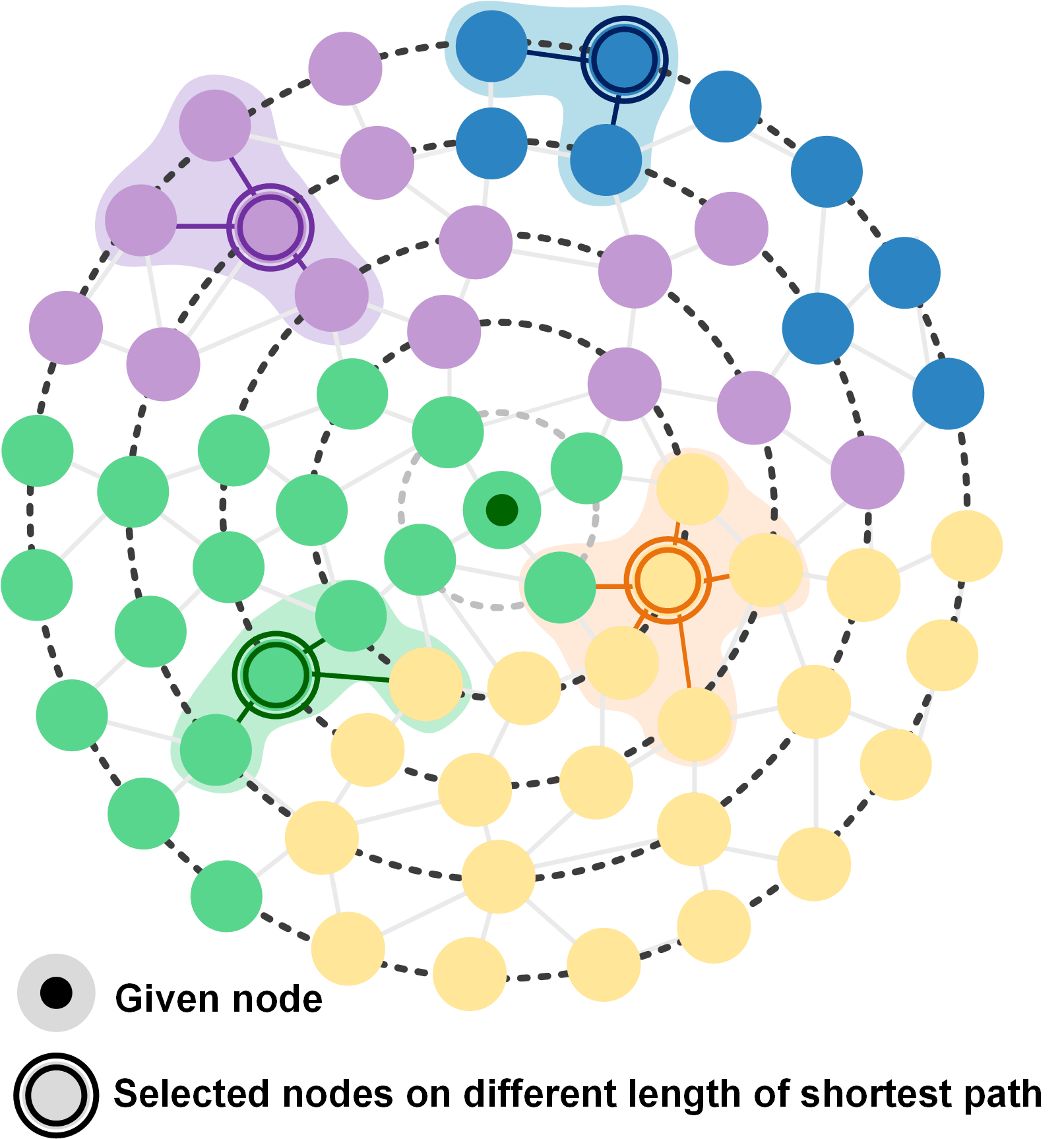} % Reduce the figure size so that it is slightly narrower than the column.
\caption{Taking the
given point as the circle's centre, different path lengths form concentric circles with different radii. All reachable nodes $V_{r}$ are divided into different sets $N_{l}$ based on the path length. We choose a random point in the node-set $N_{l}$ of each length for starting $l = 2$, and the selected nodes and their first-order adjacent nodes become the candidates set $S_i$}
\label{fig:Spath}
\end{figure}

Intuitively, good negative samples of a node should have different semantics while containing as complete a knowledge of the whole graph as possible. 
We hope the selected negative samples each belong to different communities and all communities are represented by negative samples.  
We first compute the shortest path lengths from the given node $i$ to all reachable nodes $V_{r}$. The distance from the end of the shortest path to $i$ is the path length $l$. Therefore, all reachable nodes $V_{r}$ are divided into different sets $N_{l}$ based on the path length:
\begin{equation}
    V_{r} = \{N_{l} \}_{l=2}^{L} ,
\label{eq:Spath}	
\end{equation} 
where each node of $N_{l}$ has the same distance to $i$. Taking the given point as the circle's centre, different path lengths form concentric circles with different radii, as shown in Fig.\ref{fig:Spath}. The reason for starting $l =2 $ is that negative samples should not include the first-order nearest neighbours of $i$, i.e., those nodes where $l =1 $. Next, we choose a random point in the node-set $N_{l}$ of each length, and the selected nodes and their first-order adjacent nodes become the candidate set $S_i$, as shown in all shaded areas in Fig.\ref{fig:Spath}. Finally, we select diverse negative samples from this candidate set using the same DPP idea as outlined above. 

To summarise, the first-order neighbours of nodes selected from different shortest paths can supply sufficient information from many communities but at a much smaller size. Although only first-order neighbours are collected here, collecting a higher order may further increase approximation performance; however, it will also increase the computational cost. The full procedure is summarised in Algorithm \ref{alg:spath-dpp}.

\begin{algorithm}[!t]
\caption{Shortest-path-based diverse negative sampling}
\label{alg:spath-dpp}
\SetAlgoLined
\KwIn{A graph $G$, sample length $L$}
\KwOut{$\overline{\mathcal{N}}(i)$ for all $i \in G$}
Let $\overline{\mathcal{N}} = \bf{0}$\;
\For{each node $i$}{
    Compute the shortest path lengths from $i$ to all reachable nodes $V_{r}$\;
    Divide $V_{r}$ into different sets $N_{l}$ based on the path length\;
    Let $S_i=[]$\;
    \For{l in range (2,L)}{
    Choose a random point $j$ in $N_{l}$ at each length\;
    Collect first-order neighbors $\mathcal{N}(j)$ of $j$\;
    Expand $S_i = [S_i, \mathcal{N}(j)]$\;
    }
    Compute community feature $cf_i$ using Eq.\ref{eq:communityfea}\;
    Compute candidate set feature $sf_i$ using Eq.\ref{eq:condidatefea}\;
    Compute $\mathbb{L}_{G}$ using Eq.\ref{eq:finalDPP}\;
    Define a distribution on all possible subsets $\mathcal{P}_{\mathbb{L}}(Y_{G})$ using Eq.\ref{eq:py}\;
    Obtain a sample of $\mathcal{P}_{\mathbb{L}}(Y_{G})$ using Algorithm \ref{alg:sampling_DPP}\;
    Save nodes in the sample as $\overline{\mathcal{N}}(i)$\;
}

\end{algorithm}

\subsection{GCN boosted by shortest-path-based diverse negative samples}

The classical GCN is only based on positive samples, as shown in Eq.\ref{eq:gcn}, which will inevitably lead to over-smoothing issues. With the negative samples from Algorithms  \ref{alg:dpp} or \ref{alg:spath-dpp}, we propose the following new graph convolutional operation as follows:
\begin{equation}
\begin{aligned}
	{x_i}^{(l)} =& \sum_{j \in \mathcal{N}_i \cup \{i\}}\frac{1}{\sqrt{\text{deg}(i)}\cdot\sqrt{\text{deg}(j)}}\big(\Theta^{(l)} \cdot x_j^{(l-1)}\big)\\
	&- \omega \sum_{\bar{j} \in \overline{{\mathcal{N}}_i}}\frac{1}{\sqrt{\text{deg}(i)}\cdot\sqrt{\text{deg}(\bar{j})}}\big(\Theta^{(l)} \cdot x_{\bar{j}}^{(l-1)}\big),
\end{aligned}
\label{eq:dgcn}	
\end{equation} 
where $\overline{{\mathcal{N}}_i}$ is the negative samples of node $i$, and $\omega$ is a hyper-parameter to balance the contribution of the negative samples. It is also interesting to consider that all these negative samples form a virtual graph with the same nodes as before but with negative links between them. When using the message-passing framework for mode learning, there are two messages from each node: one is positive from the neighbouring nodes, and the other is negative from the negative samples. The positive messages push all the nodes with the same semantics to have similar representations, while the negative messages push all nodes with different semantics to have different representations. This strategy is similar to clustering, where samples within the same cluster are a small distance apart, and large distances between samples indicate the sample is sitting in a different cluster. We believe that these negative messages are precisely what is missing from GCNs and are a significant element in their advancement.

The final GCN boosted by shortest-path-based diverse negative samples (SDGCN) with $L$ layers, is given in Algorithm \ref{alg:SDGCN}. Excluding negative sampling, the computational cost is double that of the original GCN. Note that although GCNs are used as the base model here, this idea can also be easily applied to other GNNs.

\begin{algorithm}[!t]
\caption{SDGCN}
\label{alg:SDGCN}
\SetAlgoLined
\KwIn{A graph $G$, sample length $L$}
\KwOut{$x^{(L)}_i$ for all $i \in G$}
\For{each level $l$}{
    \For{each node $i$ in $l$}{
        Find negative samples for $i$ using Algorithm \ref{alg:spath-dpp}\;
        Update node representation ${x_i}^{(l)}$ using Eq. \ref{eq:dgcn};
    }
}
\end{algorithm}

\begin{table*}[th]
\centering
\caption{DATASET STATISTICS}
\label{tab:dataset}
\begin{tabular}{@{}c|ccccccc@{}}
\toprule
Dataset &
  Nodes &
  Edges &
  Classes &
  Features &
  \begin{tabular}[c]{@{}c@{}}Nodes of \\ Max Subgraph\end{tabular} &
  \begin{tabular}[c]{@{}c@{}}Edges of \\ Max Subgraph\end{tabular} &
  Avg. Degree \\ \midrule
Citeseer & 3,327  & 9,104  & 6 & 3,703 & 2120   & 7358   & 2.74 \\
Cora     & 2,708  & 10,556 & 7 & 1,443 & 2485   & 10,138 & 3.90 \\
PubMed   & 19,717 & 88,648 & 3 & 500   & 19,717 & 88,648 & 4.50 \\ \bottomrule
\end{tabular}
\end{table*}

\section{Experiments}
In this section, we perform an experimental evaluation of our proposed model. We first present the data set used, our model setup, and the hyperparameter selection strategy. Then, we present and discuss the results of our experiments.

\subsection{Experiment setup}
\subsubsection{Datasets}
The datasets used are benchmark graph datasets in the citation network: Citeseer, Cora and PubMed \cite{sen2008collective}. The datasets include sparse bag-of-words feature vectors for each document as well as a list of document-to-document citation connections. Datasets are downloaded from PyTorch geometric\footnote{https://pytorch-geometric.readthedocs.io/en/latest/modules/\\datasets.html}. The dataset splitting was strictly in accordance with Kipf \& Welling\cite{DBLP:conf/iclr/KipfW17}. All graph data are undirected graphs. To better perform the experiments using the shortest path algorithm, the maximum connected subgraph of each graph data was used.
Relevant statistics on the datasets are shown in Tab.\ref{tab:dataset}. The datasets were split strictly in accordance with \cite{DBLP:conf/iclr/KipfW17}.

\begin{table*}[h]
\centering
\caption{Accuracy and MAD of all 4-layer models on three datasets}
\label{tab:ACCMAD}
\begin{tabular}{@{}c|ccc|ccc@{}}
\toprule
\multirow{2}{*}{Model} & \multicolumn{3}{c|}{Accuracy}         & \multicolumn{3}{c}{MAD}               \\ \cmidrule(l){2-7} 
                       & Citeseer    & Cora       & PubMed     & Citeseer   & Cora       & PubMed      \\ \midrule
GCN                    & 50.38$\pm$4.74  & 63.36$\pm$7.75 & 72.22$\pm$3.99 & 68.09$\pm$7.98 & 70.27$\pm$7.55 & 84.18$\pm$9.60  \\
GraphSAGE              & 62.07$\pm$5.25  & 72.90$\pm$2.65 & 74.20$\pm$2.11 & 75.93$\pm$5.47 & 70.92$\pm$5.06 & 86.04$\pm$8.32  \\
GATv2                  & 56.31$\pm$6.37  & 68.45$\pm$5.36 & 73.91$\pm$2.77 & 77.68$\pm$7.92 & 74.68$\pm$6.46 & 79.81$\pm$11.63 \\ \midrule
RGCN                   & 51.67$\pm$10.35 & 53.60$\pm$7.90 & 70.53$\pm$4.35 & 72.62$\pm$6.72 & 66.56$\pm$6.17 & 71.44$\pm$6.90  \\
MCGCN                  & 48.16$\pm$5.87  & 68.23$\pm$4.81 & 71.44$\pm$4.09 & 58.37$\pm$6.84 & 69.41$\pm$4.11 & 77.33$\pm$6.36  \\
PGCN                   & 54.84$\pm$8.46  & 58.20$\pm$8.47 & 70.42$\pm$4.58 & 74.38$\pm$8.28 & 66.85$\pm$7.07 & 77.03$\pm$6.52  \\ 
D2GCN                   & 61.66$\pm$2.66  & 70.73$\pm$4.47 & 76.11$\pm$1.13 & 73.10$\pm$6.15 & 74.26$\pm$6.39 & 89.68$\pm$4.03  \\\midrule
MAD                    & 54.28$\pm$7.53 &  65.85$\pm$8.21 & 70.48$\pm$4.29          & 80.20$\pm$4.13 & 77.28$\pm$6.05          & 81.86$\pm$8.66           \\
DGN   & 64.38$\pm$3.44          & 73.30$\pm$4.53          & 75.27$\pm$1.27         & \textbf{86.29}$\pm$\textbf{4.64} & \textbf{85.78}$\pm$\textbf{6.67} & 87.78$\pm$8.21          \\ \midrule
SDGCN & \textbf{66.58}$\pm$\textbf{3.16} & \textbf{75.49}$\pm$\textbf{1.62} & \textbf{76.87}$\pm$\textbf{1.93} & 83.37$\pm$2.38          & 81.02$\pm$3.75          & \textbf{90.43}$\pm$\textbf{3.99} \\ \bottomrule
\end{tabular}
\end{table*}

\subsubsection{Baselines}
We compare SDGCN against several other methods. The first set includes GCN and its state-of-the-art variants.
\begin{itemize}
    \item \textbf{GCN}\cite{DBLP:conf/iclr/KipfW17}: is a general GCN model introducing the traditional concept of convolution from Euclidean data to graph data with a message passing mechanism. It is also our base model.
    \item \textbf{GraphSAGE}\cite{hamilton2017inductive}: is a general framework for generating node embedding by sampling and aggregating features from the neighbourhood of a node.
    \item \textbf{GATv2}\cite{brody2021attentive}: is a dynamic graph attention variant that is strictly more expressive than the graph attention network. 
\end{itemize}
The second set includes the only three negative sampling methods available to date in the GCNs domain. Note that these methods only use the results of negative sampling in the loss function. Thus, we first get negative samples using these methods, then put the selected samples into the convolution operation using Eq.\ref{eq:dgcn}. We further compare our previous work D2GCN.
\begin{itemize}
    \item \textbf{RGCN}\cite{kim2021find}:select negative samples in a purely random way.
    \item \textbf{MCGCN}\cite{yang2020understanding}: selects negative samples based on Monte Carlo chains.
    \item \textbf{PGCN} \cite{ying2018graph}: is based on  personalised PageRank.
    \item \textbf{D2GCN}\cite{D2DCN}: uses only the node representations to calculate $\mathbb{L}$-ensemble and uses depth-first search based method to get the candidate sets.
\end{itemize}
The third set includes two latest methods for
over-smoothing.
\begin{itemize}
    \item \textbf{MAD}\cite{chen2020measuring}: proposes a quantitative metric Mean Average Distance (MAD) and uses a MADGap-based regularizer to alleviate over-smoothing.
    \item \textbf{DGN}\cite{DBLP:conf/nips/Zhou0LZCH20}: introduces differentiable group normalization (DGN). It normalizes nodes within the same group independently to increase their smoothness and separates node distributions among different groups.
\end{itemize}
To ensure the consistency and fairness of the experiments, all the graph convolution models have the same structure and are initialised and trained with the same methods. All models run our own implementation.

\subsubsection{Experiment Settings}

The experimental task was standard node classification.
We set the maximum length of the shortest path $L$ to 6, which means after throwing away the first-order nearest neighbours, there are still 5 nodes on the end of the different shortest path. For  Citeseer and Cora, because of their relatively small number of nodes, we perform negative sampling on all nodes. For Pubmed, we only perform negative sampling on nodes with a degree greater than 1 (about 50\%). The negative rate is a trainable parameter and trained all models with $4$ layers. Each model was trained for 200 epochs on Cora and Citeseer and for 100 epochs with PubMed, using an Adam optimiser with a learning rate of 0.01. Tests for each model with each dataset were conducted 10 times. All experiments were conducted on an Intel(R) Xeon(R) CPU @ 2.00GHz and NVIDIA Tesla T4 GPU. The code was implemented in PyTorch.\footnote{The code is available at https://github.com/Wei9711/NegGCNs.}

\subsubsection{Metrics}

It was our goal to verify two capabilities of the proposed model: one is the ability to improve the prediction performance, and the other is to alleviate the over-smoothing problem. Hence, we used the following two metrics:

\begin{itemize}
\item \textbf{Accuracy} is calculated by dividing the number of correctly classified nodes by the total number of test nodes. (The larger, the better);
\item \textbf{Mean Average Distance
(MAD)} \cite{chen2020measuring} reflects the smoothness of graph representation (The larger, the better):
\begin{equation}
\begin{aligned}
	\text{MAD} =\frac{\sum_i D_i}{\sum_i 1\left(D_i\right)}, \quad D_i=\frac{\sum_j D_{i j}}{\sum_j 1\left(D_{i j}\right)} ,
\end{aligned}
\end{equation} 
where $D_{ij} = 1 - \cos(x_i, x_j)$ is the cosine distance between nodes $i$ and $j$.
\end{itemize}

\subsection{Experiment Results}

\subsubsection{Comparison to different models}
We compare SDGCN against several different methods. The results of all models with 4 layers on three datasets are shown in Tab.\ref{tab:ACCMAD}. In relation to accuracy, our model was marginally better than the other models. For MAD,  our model also achieves superior performance and is only lower than the DGN method on the Cora and Citeseer datasets. 

Specifically, SDGCN achieves better performance over the current state-of-the-art.
GCN variant: GraphSAGE and GATv2. Compared with these two methods, SDGCN not only involves neighbouring nodes(positive samples) in message passing but also integrates the negative samples obtained by our method. Thus, it achieves more satisfactory results in the 4-layer network. 

Compared with the three negative sampling methods available to date in the GCN domain (i.e., RGCN, MCGCN, PGCN), although these methods add negative samples into the convolution operation, they cannot yield consistent results, and their performance fluctuates greatly across different datasets. This observation suggests that adding negative samples to the convolution reduces the over-smoothing to some extent, but choosing the appropriate negative samples is not trivial. Additional effort needs to be given to the procedure for selecting appropriate negative samples. Since SDGCN not only considers node features but also incorporates graph structure information when computing $\mathbb{L}$-ensemble, its results are better than our previous method D2GCN.

Compared with the two latest methods for over-smoothing (i.e., MAD, DGN), our model also achieves satisfactory results for accuracy. Although for MAD, our model is not as good as DGN on the Citeseer and Cora. These two outperform the other methods by a wide margin, which verifies that alleviating the over-smoothing problem will, in turn, improve prediction accuracy. Moreover, in addition to MAD (MADGap-based regularizer) and DGN (differentiable group normalization), our approach proposes a novel idea to solve the over-smoothing problem from the perspective of negative samples, which aggregates differentiated information for each node in the process of message-passing.

\subsubsection{Hyperparameters Sensitivity}
\begin{figure*}[ht]
\centering
\includegraphics[width=\textwidth]{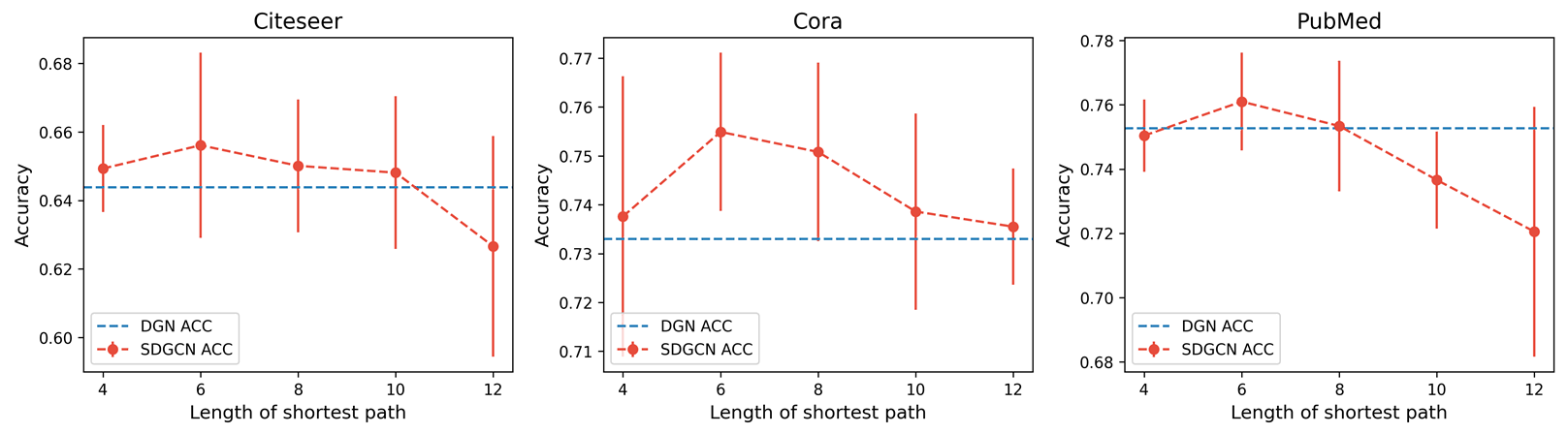}
 % Reduce the figure size so that it is slightly narrower than the column. Don't use precise values for figure width.This setup will avoid overfull boxes.
\caption{The variation in the length of the shortest path and the corresponding Accuracy on three datasets.}
\label{fig:different Spath}
\end{figure*}

\begin{figure*}[h]
\centering
\includegraphics[width=0.85\textwidth]{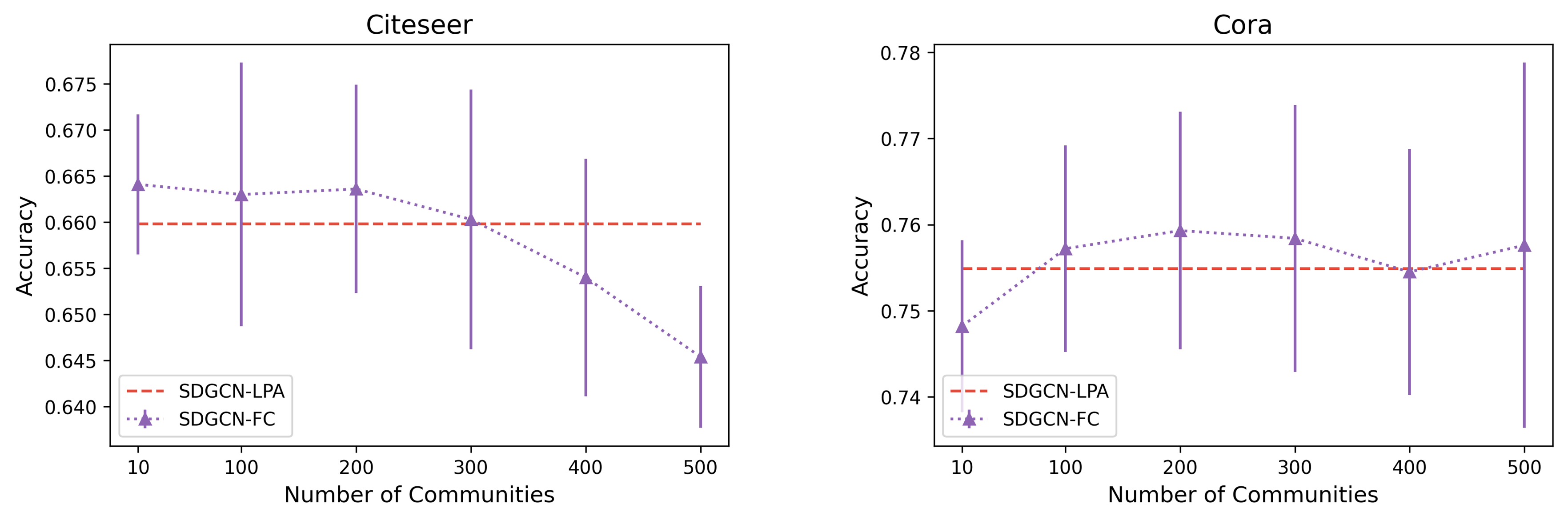}% Reduce the figure size so that it is slightly narrower than the column.
\caption{The variation in the number of communities getting by FC method and the corresponding Accuracy on Citeseer and Cora. The SDGCN-LPA is the reference line.}
\label{fig:DiffCom}
\end{figure*}

\begin{figure*}[ht]
\centering
\includegraphics[width=0.85\textwidth]{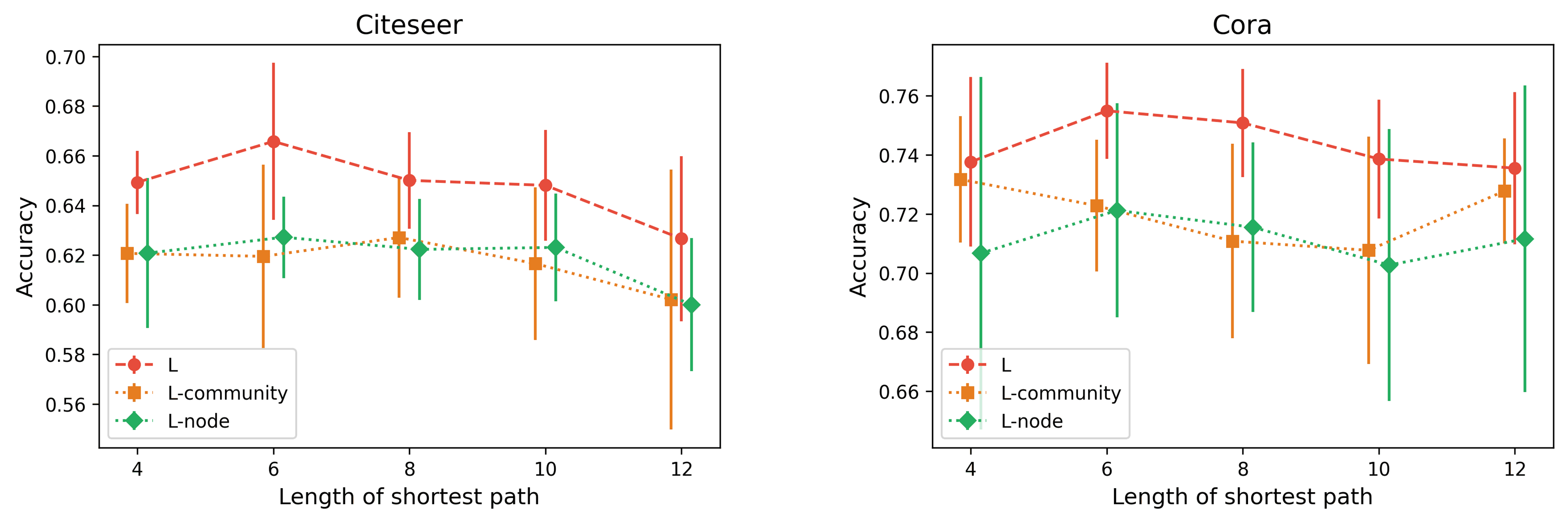} % Reduce the figure size so that it is slightly narrower than the column.
\caption{The variation in the length of the shortest path with different $\mathbb{L}$ construction methods and the corresponding accuracy in Citeseer and Cora.}
\label{fig:L-Acc}
\end{figure*}

\begin{figure*}[t]
\centering
\includegraphics[width=\textwidth]{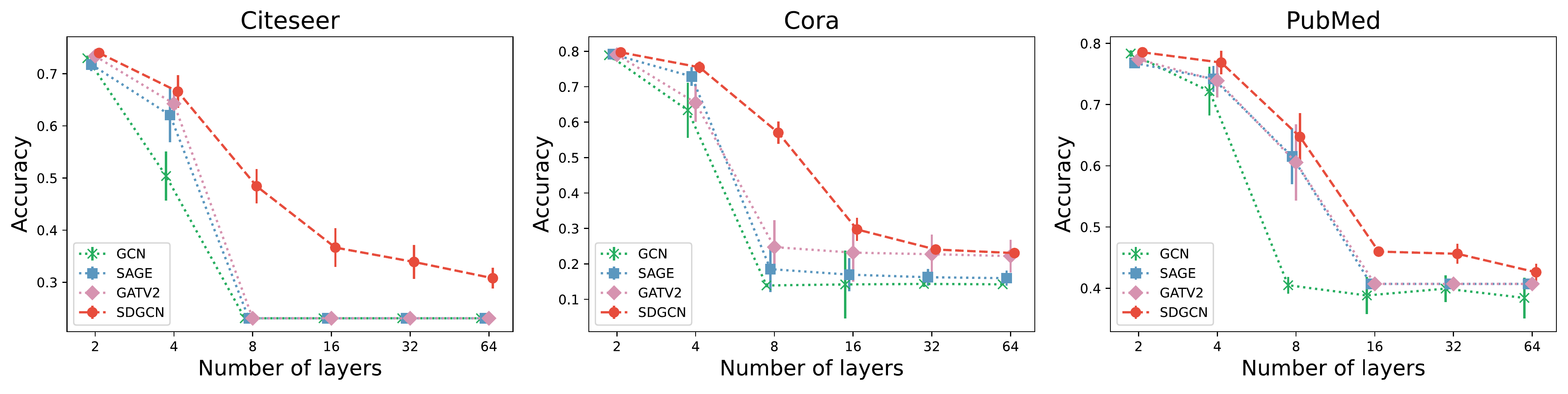}
\caption{The variation in the number of layers and the corresponding Accuracy on three datasets.}
\label{fig:deeperLayer}
\end{figure*}

In this section, we examine the sensitivity of SDGCN to the shortest path length L. Fig.\ref{fig:different Spath} shows the variation in the length of the shortest path and the corresponding accuracy on three datasets. We chose the DGN, which obtained outstanding results as the reference line. As can be seen in Fig. \ref{fig:different Spath}, our method is insensitive to the choice of the shortest path length, especially for the Cora dataset, where the results for different settings exceed DGN. An interesting finding is that after the shortest path length is greater than 6, as the length increases, which means the number of candidate sets becomes larger, the accuracy decreases to some extent on all datasets. One conjecture is that those nodes in a graph with the greatest variance are relatively constant, and as the number of candidate nodes in the set increases, the probability that the DPP always picks these with large variance increases. This results in the selected negative sample not containing enough differentiated information, which leads to a decrease in accuracy.

\subsubsection{Different methods to get communities} To test whether our method is robust to the community detection results, we choose an alternative method - Fluid Communities (FC)\cite{DBLP:conf/complexnetworks/ParesGVMALCS17}. We further test the different numbers of communities obtained by FC, and the number of communities is set to \{10, 100, 200, 300, 400, 500\}. We compared it with LPA's\cite{DBLP:journals/corr/abs-1103-4550} method on the Citeseer and Cora datasets. Fig.\ref{fig:DiffCom} show that even though we change the method to get communities, our SDGCN still can get satisfying results. For Cora (the number of communities ranges from 100 to 500) and Citeseer (the number of communities ranges from 10 to 300), SDGCN-FC obtains the same outstanding results as SDGNC-LPA. Regardless of the method used to obtain communities, the accuracy of SDGCN is stable.

\subsubsection{Ablation study for $\mathbb{L}$-ensemble}
The key to utilizing DPP is in defining an $\mathbb{L}$-ensemble. In the previous comparison experiments, we used the graph's community information and the node features to construct the $\mathbb{L}$-ensemble. In this section, using two different methods, we examine the effect of different $\mathbb{L}$ construction methods on accuracy in Citeseer and Cora.

The first method, only uses \textbf{community information} to construct the $\mathbb{L}$-ensemble, named $\mathbb{L}_{community}$. For a given node $i$, the quality term in Eq.\ref{eq:quality} becomes 
\begin{equation}
    \begin{aligned}
        &q_{i,j} =  \cos(a_{i},a_{j}),
        \\
        &{q_{i,j^{\prime}}} = \cos(a_{i},{a_{j^{\prime}}}),
    \end{aligned}
\label{eq:quality-LC}	
\end{equation}
where $a_{i}, a_{j},{a_{j^{\prime}}}$ represents the feature expression of the node belonging to its community. The diversity term in Eq.\ref{eq:diversity} becomes
\begin{equation}
    {\phi_{j}^{T}\phi_{j^{\prime}}}= \cos({x_{j}},{a_{j^{\prime}}})\cos({a_{j}},{x_{j^{\prime}}}),
\label{eq:diversity-LC}	
\end{equation}
where $j,{j^{\prime}} \in S_i$. Then, we put Eq.\ref{eq:quality-LN} and Eq.\ref{eq:diversity-LN} into Eq.\ref{eq:decomDpp} to get $\mathbb{L}_{node}$.

The second method, keeps the \textbf{node features} to construct the $\mathbb{L}$-ensemble, named $\mathbb{L}_{node}$. For a given node $i$, the quality term in Eq.\ref{eq:quality} becomes 
\begin{equation}
    \begin{aligned}
    &q_{i,j} = \cos(a_{i},b_{i}),
    \\
    &{q_{i,j^{\prime}}} = \cos(a_{i},b_{i}),
    \end{aligned}
\label{eq:quality-LN}	
\end{equation}
where $a_{i}$ represents the community feature of the given node and $b_{i}$ represents the features of candidate set $S_i$. The diversity term in Eq.\ref{eq:diversity} becomes
\begin{equation}
    {\phi_{j}^{T}\phi_{j^{\prime}}}= \exp{(\cos(({x_j},{x_{j^{\prime}}})-1))},
\label{eq:diversity-LN}	
\end{equation}
where $j,{j^{\prime}} \in S_i$. Then, we put Eq.\ref{eq:quality-LN} and Eq.\ref{eq:diversity-LN} into Eq.\ref{eq:decomDpp} to get $\mathbb{L}_{node}$.

The results of using different $\mathbb{L}$-ensembles on Citeseer and Cora are shown in Fig.\ref{fig:L-Acc}. We can see that both the $\mathbb{L}_{community}$ and $\mathbb{L}_{node}$ methods perform very similarly for different shortest path lengths, and neither is as good as $\mathbb{L}$ which uses both structural and node information. Another noteworthy finding is that the results obtained using the $\mathbb{L}$-matrix have a much smaller standard deviation than the other two methods. This indicates that incorporating both node features and graph structure information allows DPP to select quality negative samples more consistently.

% \begin{figure}[ht]
% \centering
% \includegraphics[width=0.9\columnwidth]{fig/Cora Acc-L.png}% Reduce the figure size so that it is slightly narrower than the column.
% \caption{The variation in the length of the shortest path with different $\mathbb{L}$ construction methods and the corresponding accuracy in Cora}
% \label{fig:CoraL}
% \end{figure}

\subsubsection{Deep-layer 
study}
We further compared SDGCN to GCN \cite{DBLP:conf/iclr/KipfW17}, GraphSAGE\cite{hamilton2017inductive}, and GATv2 \cite{brody2021attentive} on three datasets using 2, 4, 8, 16, 32, and 64 layers. The results are presented in Fig. \ref{fig:deeperLayer}. All models are trained for 200 epochs with two layers and 400 epochs for other layer settings. Experiments were conducted ten times. We perform negative sampling on the top 10\% of nodes with the highest degrees for training efficiency. Our proposed model consistently achieved the highest accuracy across all datasets. Although our model's performance decreased with increasing depth, the decline rate was significantly slower than the other three models, especially at the 8-layer. This observation confirms that our approach effectively mitigates the over-smoothing issue and consequently enhances prediction accuracy.

\begin{figure}[th]
\centering
\includegraphics[width=0.45\columnwidth]{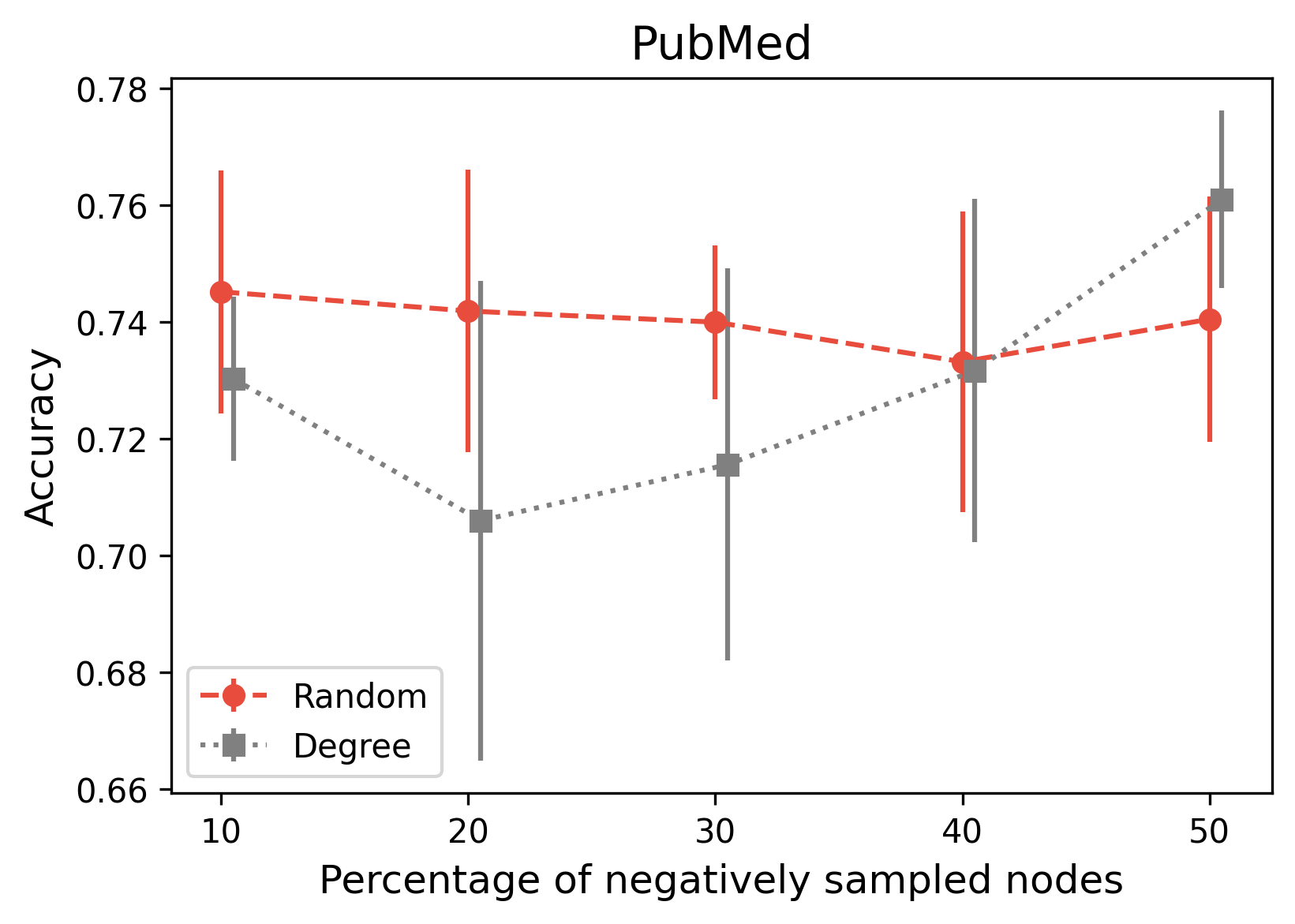} % Reduce the figure size so that it is slightly narrower than the column.
\caption{The effect of choosing a different number of nodes for negative sampling on accuracy in PubMed.}
\label{fig:Pubmed}
\end{figure}

\subsubsection{Number of negative samples}
For PubMed, we only perform negative sampling on nodes with a degree greater than 1 (about 50\%). In this section, we further explore the effect of the number of nodes that perform negative sampling on the SDGCN. The number of selected nodes is set to \{10\%, 20\%, 30\%, 40\%, 50\%\}.
We compared two selection methods, one for purely random selection and the other for selecting those with the largest node degree. From Fig.\ref{fig:Pubmed}, it can be seen that when the number of selected nodes is less than 40\%, the result for the random selection method is better than the method of selecting the node with the maximum degree. A reasonable inference is that not all nodes with a large degree are "important" to the 
graph. Thus, when the number of selected nodes is small, those nodes with the largest degree are relatively constant, while randomly selected nodes may contain differentiated information in the graph, which leads to better results. When the number of selected nodes is greater than 40\%, selecting the nodes with a large degree is better than random selection. There are 9094 nodes with a degree of 1 in PubMed, which is about 50\% of the total number of nodes. Nodes with a degree of 1 can be considered as margins of the graph. We consider these nodes to be less important to the graph, so selecting all points with a node degree greater than 1 for negative sampling is more effective than randomly selecting 50\% of the nodes.

\section{Conclusion}
In this paper, we utilize quality-diversity decomposition in DPP to incorporate graph structure information in $\mathbb{L}$-ensemble and select negative samples to boost graph convolutional neural networks. To the best of our knowledge, we are the first to introduce decomposed DPP to GCNs for negative sampling. We further presented a shortest-path-based heuristic approximation method to reduce the computational cost greatly. The experiment evaluations show that the proposed SDGCN consistently performs better than alternative methods. In addition to greater predictive accuracy, the method also helps to prevent over-smoothing. With this study, we identify that negative samples are important to graph neural networks and should be considered in future works. Note that the proposed idea can be applied to other graph neural networks besides GCN.

In future research, we will continue to investigate how to speed up the DPP sampling process in the algorithm. Another interesting follow-up work would be to investigate a more effective aggregation of positive and negative samples, as the current solution may lose some information when summing the samples together, especially when the samples are diverse.

\section*{Acknowledgments}
This work is supported by the Australian Research Council under Australian Laureate Fellowships FL190100149 and Discovery Early Career Researcher Award DE200100245.

%Bibliography
\bibliographystyle{unsrt}  
\bibliography{references}

\begin{thebibliography}{10}

\bibitem{lecun1995convolutional}
Yann LeCun, Yoshua Bengio, et~al.
\newblock Convolutional networks for images, speech, and time series.
\newblock {\em The handbook of brain theory and neural networks},
  3361(10):1995, 1995.

\bibitem{DBLP:journals/tkde/TrotzekKF20}
Marcel Trotzek, Sven Koitka, and Christoph~M. Friedrich.
\newblock Utilizing neural networks and linguistic metadata for early detection
  of depression indications in text sequences.
\newblock {\em IEEE Transactions on Knowledge and Data Engineering},
  32(3):588--601, 2020.

\bibitem{duan2021refined}
Wei Duan, Yufei Chen, Qi~Zhang, Xiang Lin, and Xiaoyu Yang.
\newblock Refined tooth and pulp segmentation using u-net in cbct image.
\newblock {\em Dentomaxillofacial Radiology}, 50(6):20200251, Sep 2021.

\bibitem{DBLP:conf/icassp/ChoiFSC17}
Keunwoo Choi, Gy{\"{o}}rgy Fazekas, Mark~B. Sandler, and Kyunghyun Cho.
\newblock Convolutional recurrent neural networks for music classification.
\newblock In {\em 2017 {IEEE} International Conference on Acoustics, Speech and
  Signal Processing{(ICASSP)}, New Orleans, LA, USA,}, pages 2392--2396, 2017.

\bibitem{DBLP:journals/tnn/ChenWLXW18}
Xin Chen, Jian Weng, Wei Lu, Jiaming Xu, and Jia{-}Si Weng.
\newblock Deep manifold learning combined with convolutional neural networks
  for action recognition.
\newblock {\em IEEE Transactions on Neural Networks and Learning Systems},
  29(9):3938--3952, 2018.

\bibitem{chakrabarti2006graph}
Deepayan Chakrabarti and Christos Faloutsos.
\newblock Graph mining: laws, generators, and algorithms.
\newblock {\em ACM Computing Surveys}, 38(1):1--69, 2006.

\bibitem{DBLP:conf/iclr/XuHLJ19}
Keyulu Xu, Weihua Hu, Jure Leskovec, and Stefanie Jegelka.
\newblock How powerful are graph neural networks?
\newblock In {\em Proceedings of the 7th International Conference on Learning
  Representations ({ICLR}), New Orleans, LA, US}, 2019.

\bibitem{DBLP:journals/tnn/WuPCLZY21}
Zonghan Wu, Shirui Pan, Fengwen Chen, Guodong Long, Chengqi Zhang, and
  Philip~S. Yu.
\newblock A comprehensive survey on graph neural networks.
\newblock {\em IEEE Transactions on Neural Networks and Learning Systems},
  32(1):4--24, 2021.

\bibitem{DBLP:journals/corr/LiTBZ15}
Yujia Li, Daniel Tarlow, Marc Brockschmidt, and Richard~S. Zemel.
\newblock Gated graph sequence neural networks.
\newblock In {\em Proceedings of the 4th International Conference on Learning
  Representations ({ICLR}), San Juan, Puerto Rico}, 2016.

\bibitem{DBLP:conf/icml/DaiKDSS18}
Hanjun Dai, Zornitsa Kozareva, Bo~Dai, Alexander~J. Smola, and Le~Song.
\newblock Learning steady-states of iterative algorithms over graphs.
\newblock In {\em Proceedings of the 35th International Conference on Machine
  Learning({ICML}), Stockholmsm{\"{a}}ssan, Stockholm, Sweden}, pages
  1114--1122, 2018.

\bibitem{DBLP:journals/corr/BrunaZSL13}
Joan Bruna, Wojciech Zaremba, Arthur Szlam, and Yann LeCun.
\newblock Spectral networks and locally connected networks on graphs.
\newblock In {\em Proceedings of the 2nd International Conference on Learning
  Representations ({ICLR}), Banff, AB, Canada}, 2014.

\bibitem{DBLP:conf/iclr/KipfW17}
Thomas~N. Kipf and Max Welling.
\newblock Semi-supervised classification with graph convolutional networks.
\newblock In {\em 5th International Conference on Learning Representations
  ({ICLR}), Toulon, France}, 2017.

\bibitem{hamilton2017inductive}
William~L. Hamilton, Zhitao Ying, and Jure Leskovec.
\newblock Inductive representation learning on large graphs.
\newblock In {\em Proceedings of the 30th International Conference on Neural
  Information Processing Systems ({NIPS}), Long Beach, CA, United States},
  pages 1024--1034, 2017.

\bibitem{9476188}
Luca Pasa, Nicolò Navarin, and Alessandro Sperduti.
\newblock Multiresolution reservoir graph neural network.
\newblock {\em IEEE Transactions on Neural Networks and Learning Systems},
  pages 1--12, 2021.

\bibitem{9576073}
Diego Valsesia, Giulia Fracastoro, and Enrico Magli.
\newblock Ran-gnns: Breaking the capacity limits of graph neural networks.
\newblock {\em IEEE Transactions on Neural Networks and Learning Systems},
  pages 1--10, 2021.

\bibitem{DBLP:conf/ijcai/YuYZ18}
Bing Yu, Haoteng Yin, and Zhanxing Zhu.
\newblock Spatio-temporal graph convolutional networks: {A} deep learning
  framework for traffic forecasting.
\newblock In {\em Proceedings of the 27th International Joint Conference on
  Artificial Intelligence ({IJCAI}), Stockholm, Sweden}, pages 3634--3640,
  2018.

\bibitem{DBLP:conf/ijcai/WuPLJZ19}
Zonghan Wu, Shirui Pan, Guodong Long, Jing Jiang, and Chengqi Zhang.
\newblock Graph wavenet for deep spatial-temporal graph modeling.
\newblock In {\em Proceedings of the Twenty-Eighth International Joint
  Conference on Artificial Intelligence ({IJCAI})}, pages 1907--1913, 2019.

\bibitem{geerts2021let}
Floris Geerts, Filip Mazowiecki, and Guillermo~A. P{\'{e}}rez.
\newblock Let's agree to degree: comparing graph convolutional networks in the
  message-passing framework.
\newblock In {\em Proceedings of the 38th International Conference on Machine
  Learning ({ICML}), Virtual Event}, pages 3640--3649, 2021.

\bibitem{chen2020measuring}
Deli Chen, Yankai Lin, Wei Li, Peng Li, Jie Zhou, and Xu~Sun.
\newblock Measuring and relieving the over-smoothing problem for graph neural
  networks from the topological view.
\newblock In {\em Proceedings of The 34th {AAAI} Conference on Artificial
  Intelligence ({AAAI}), New York, NY, USA}, pages 3438--3445, 2020.

\bibitem{kim2021find}
Dongkwan Kim and Alice~H. Oh.
\newblock How to find your friendly neighborhood: graph attention design with
  self-supervision.
\newblock In {\em Proceedings of the 9th International Conference on Learning
  Representations ({ICLR}), Virtual Event, Austria}, 2021.

\bibitem{yang2020understanding}
Zhen Yang, Ming Ding, Chang Zhou, Hongxia Yang, Jingren Zhou, and Jie Tang.
\newblock Understanding negative sampling in graph representation learning.
\newblock In {\em Proceedings of the 26th ACM SIGKDD International Conference
  on Knowledge Discovery \& Data Mining ({KDD}), Virtual Event, CA, USA}, pages
  1666--1676, 2020.

\bibitem{ying2018graph}
Rex Ying, Ruining He, Kaifeng Chen, Pong Eksombatchai, William~L. Hamilton, and
  Jure Leskovec.
\newblock Graph convolutional neural networks for web-scale recommender
  systems.
\newblock In {\em Proceedings of the 24th {ACM} {SIGKDD} International
  Conference on Knowledge Discovery {\&} Data Mining {(KDD)}, London, UK},
  pages 974--983, 2018.

\bibitem{kulesza2012determinantal}
Alex Kulesza and Ben Taskar.
\newblock Determinantal point processes for machine learning.
\newblock {\em Foundations and Trends in Machine Learning}, 5(2-3):123--286,
  2012.

\bibitem{sen2008collective}
Prithviraj Sen, Galileo Namata, Mustafa Bilgic, Lise Getoor, Brian Gallagher,
  and Tina Eliassi{-}Rad.
\newblock Collective classification in network data.
\newblock {\em {AI} Magazine}, 29(3):93--106, 2008.

\bibitem{hough2009zeros}
J.~Ben Hough, Manjunath Krishnapur, Yuval Peres, and B{\'{a}}lint Vir{\'{a}}g.
\newblock {\em Zeros of gaussian analytic functions and determinantal point
  processes}, volume~51 of {\em University Lecture Series}.
\newblock American Mathematical Society, 2009.

\bibitem{zhu2020hgcn}
Zhihua Zhu, Xinxin Fan, Xiaokai Chu, and Jingping Bi.
\newblock Hgcn: A heterogeneous graph convolutional network-based deep learning
  model toward collective classification.
\newblock In {\em Proceedings of the 26th ACM SIGKDD International Conference
  on Knowledge Discovery \& Data Mining ({KDD}), Virtual Event, CA, USA}, pages
  1161--1171, 2020.

\bibitem{yang2021interpretable}
Yaming Yang, Ziyu Guan, Jianxin Li, Wei Zhao, Jiangtao Cui, and Quan Wang.
\newblock Interpretable and efficient heterogeneous graph convolutional
  network.
\newblock {\em IEEE Transactions on Knowledge and Data Engineering}, pages
  1--1, 2021.

\bibitem{li2021higher}
Jianxin Li, Hao Peng, Yuwei Cao, Yingtong Dou, Hekai Zhang, Philip Yu, and
  Lifang He.
\newblock Higher-order attribute-enhancing heterogeneous graph neural networks.
\newblock {\em IEEE Transactions on Knowledge and Data Engineering}, pages
  1--1, 2021.

\bibitem{weisfeiler1968reduction}
Boris Weisfeiler and Andrei Leman.
\newblock The reduction of a graph to canonical form and the algebra which
  appears therein.
\newblock {\em NTI, Series}, 2(9):12--16, 1968.

\bibitem{mnih2013learning}
Andriy Mnih and Koray Kavukcuoglu.
\newblock Learning word embeddings efficiently with noise-contrastive
  estimation.
\newblock {\em Advances in neural information processing systems},
  26:2265--2273, 2013.

\bibitem{mikolov2013distributed}
Tomas Mikolov, Ilya Sutskever, Kai Chen, Greg Corrado, and Jeffrey Dean.
\newblock Distributed representations of words and phrases and their
  compositionality.
\newblock {\em arXiv preprint arXiv:1310.4546}, 2013.

\bibitem{kang2013fast}
Byungkon Kang.
\newblock Fast determinantal point process sampling with application to
  clustering.
\newblock In {\em Proceedings of 27th Annual Conference on Neural Information
  Processing Systems ({NIPS}), Lake Tahoe, Nevada, United States}, pages
  2319--2327, 2013.

\bibitem{gillenwater2012near}
Jennifer Gillenwater, Alex Kulesza, and Ben Taskar.
\newblock Near-optimal {MAP} inference for determinantal point processes.
\newblock In {\em Proceedings of 26th Annual Conference on Neural Information
  Processing Systems ({NIPS}), Lake Tahoe, Nevada, United States}, pages
  2744--2752, 2012.

\bibitem{affandi2014learning}
Raja~Hafiz Affandi, Emily~B. Fox, Ryan~P. Adams, and Benjamin Taskar.
\newblock Learning the parameters of determinantal point process kernels.
\newblock In {\em Proceedings of the 31th International Conference on Machine
  Learning ({ICML}), Beijing, China}, pages 1224--1232, 2014.

\bibitem{gillenwater2014expectation}
Jennifer Gillenwater, Alex Kulesza, Emily~B. Fox, and Benjamin Taskar.
\newblock Expectation-maximization for learning determinantal point processes.
\newblock In {\em Proceedings of 27th Annual Conference on Neural Information
  Processing Systems ({NIPS}), Montreal, Quebec, Canada}, pages 3149--3157,
  2014.

\bibitem{gillenwater2012discovering}
Jennifer Gillenwater, Alex Kulesza, and Ben Taskar.
\newblock Discovering diverse and salient threads in document collections.
\newblock In {\em Proceedings of the 2012 Joint Conference on Empirical Methods
  in Natural Language Processing and Computational Natural Language Learning
  ({EMNLP-CoNLL}), Jeju Island, Korea}, pages 710--720, 2012.

\bibitem{affandi2012markov}
Raja~Hafiz Affandi, Alex Kulesza, and Emily~B. Fox.
\newblock Markov determinantal point processes.
\newblock In {\em Proceedings of the 28th Conference on Uncertainty in
  Artificial Intelligence ({UAI}), Catalina Island, CA, USA}, pages 26--35,
  2012.

\bibitem{snoek2013determinantal}
Jasper Snoek, Richard~S. Zemel, and Ryan~Prescott Adams.
\newblock A determinantal point process latent variable model for inhibition in
  neural spiking data.
\newblock In {\em Proceedings of the 27th Annual Conference on Neural
  Information Processing Systems ({NIPS}), Lake Tahoe, Nevada, United States},
  pages 1932--1940, 2013.

\bibitem{qiao2015diversified}
Maoying Qiao, Wei Bian, Richard~Yi Da~Xu, and Dacheng Tao.
\newblock Diversified hidden models for sequential labeling.
\newblock {\em IEEE Transactions on Knowledge and Data Engineering},
  27(11):2947--2960, 2015.

\bibitem{DBLP:conf/aaai/YaoFZWCX16}
Jin{-}ge Yao, Feifan Fan, Wayne~Xin Zhao, Xiaojun Wan, Edward~Y. Chang, and
  Jianguo Xiao.
\newblock Tweet timeline generation with determinantal point processes.
\newblock In {\em Proceedings of the Thirtieth {AAAI} Conference on Artificial
  Intelligence {(AAAI)}, Phoenix, Arizona, {USA}}, pages 3080--3086, 2016.

\bibitem{DBLP:conf/acl/ChoLFL19}
Sangwoo Cho, Logan Lebanoff, Hassan Foroosh, and Fei Liu.
\newblock Improving the similarity measure of determinantal point processes for
  extractive multi-document summarization.
\newblock In {\em Proceedings of the 57th Conference of the Association for
  Computational Linguistics {(ACL)} Florence, Italy, Volume 1: Long Papers},
  pages 1027--1038, 2019.

\bibitem{DBLP:conf/ijcai/ZhengL20}
Jiping Zheng and Ganfeng Lu.
\newblock k-sdpp: Fixed-size video summarization via sequential determinantal
  point processes.
\newblock In Christian Bessiere, editor, {\em Proceedings of the Twenty-Ninth
  International Joint Conference on Artificial Intelligence {(IJCAI)}}, pages
  774--781, 2020.

\bibitem{D2DCN}
Wei Duan, Junyu Xuan, Wayne~Xin Zhao, Maoying Qiao, and Jie Lu.
\newblock Learning from the dark: Boosting graph convolutional neural networks
  with diverse negative samples.
\newblock In {\em Proceedings of the Thirty-Sixth {AAAI} Conference on
  Artificial Intelligence{(AAAI)}, February 22-March 1, 2022, Vancouver, BC,
  {Canada}}, Accepted.

\bibitem{DBLP:journals/corr/abs-1103-4550}
Gennaro Cordasco and Luisa Gargano.
\newblock Community detection via semi-synchronous label propagation
  algorithms.
\newblock {\em CoRR}, abs/1103.4550, 2011.

\bibitem{hough2006determinantal}
J~Ben Hough, Manjunath Krishnapur, Yuval Peres, and B{\'a}lint Vir{\'a}g.
\newblock Determinantal processes and independence.
\newblock {\em Probability Surveys}, 3:206--229, 2006.

\bibitem{brody2021attentive}
Shaked Brody, Uri Alon, and Eran Yahav.
\newblock How attentive are graph attention networks?, 2021.

\bibitem{DBLP:conf/nips/Zhou0LZCH20}
Kaixiong Zhou, Xiao Huang, Yuening Li, Daochen Zha, Rui Chen, and Xia Hu.
\newblock Towards deeper graph neural networks with differentiable group
  normalization.
\newblock In {\em Advances in Neural Information Processing Systems 33: Annual
  Conference on Neural Information Processing Systems 2020{ (NIPS)}, virtual},
  2020.

\bibitem{DBLP:conf/complexnetworks/ParesGVMALCS17}
Ferran Par{\'{e}}s, Dario Garcia{-}Gasulla, Armand Vilalta, Jonathan Moreno,
  Eduard Ayguad{\'{e}}, Jes{\'{u}}s Labarta, Ulises Cort{\'{e}}s, and Toyotaro
  Suzumura.
\newblock Fluid communities: {A} competitive, scalable and diverse community
  detection algorithm.
\newblock In {\em Proceedings of The Sixth International Conference on Complex
  Networks and Their Applications, ({COMPLEX} {NETWORKS}), Lyon, France},
  volume 689, pages 229--240, 2017.

\end{thebibliography}

\end{document}